%% file: nestok_arxiv.tex
\documentclass{article} %
\usepackage{iclr2027_conference,times}

\input{math_commands.tex}

\usepackage{hyperref}
\usepackage{cleveref}
\crefname{figure}{figure}{figure}
\Crefname{figure}{Figure}{Figure}
\usepackage{url}

\usepackage{xspace}
\usepackage{dsfont}
\usepackage{afterpage}

\usepackage{enumitem}
\usepackage{booktabs}
\usepackage{caption} %
\usepackage{multirow} %
\usepackage{enumitem}
\usepackage{colortbl} %
\usepackage{bbm}
\usepackage{algorithm}
\usepackage{algorithmic}

\usepackage{setspace}                       %
\usepackage{lipsum} %
\usepackage{subcaption} %
\usepackage{makecell}

\usepackage{mathtools} %
\usepackage{bm} %
\usepackage{soul} %
\usepackage{nicefrac}
\usepackage{csquotes}

\usepackage{duckuments}
\usepackage{wrapfig}

\definecolor{cornellred}{rgb}{0.7, 0.11, 0.11}
\definecolor{cadmiumgreen}{rgb}{0.0, 0.42, 0.24}
\definecolor{aliceblue}{rgb}{0.91, 0.94, 0.97}
\definecolor{darkblue}{rgb}{0.83, 0.89, 0.97}
\definecolor{Red7}{rgb}{0.941, 0.243, 0.243}
\definecolor{Green7}{RGB}{55, 178, 77}
\definecolor{Blue9}{rgb}{0.098,0.3,0.9}

\usepackage{pifont}%

\sethlcolor{aliceblue}
\usepackage{amssymb}

\hypersetup{
  linkcolor = cornellred,
  citecolor  = cadmiumgreen,
  colorlinks = true,
  urlcolor = Blue9
}

\newif\ifunderreview
\underreviewfalse %

\iclrfinalcopy

\title{NesTok: Nested Self-Aligned 1D Tokenizer for Autoregressive Image Generation }

\author{Jiawei Zhang$^1$\thanks{Equal contribution: \{zjw1637, shuhao\}@ncepu.edu.cn.}{\quad}Shuhao Liu$^1$$^\ast${\quad}Rong Huang$^1${\quad}Yuancheng Li$^1$\\
\textbf{Zhihui Li$^2${\quad}Xiaojun Chang$^2${\quad}Changlin Li$^{3}$}\thanks{Corresponding author: changlinli.ai@gmail.com.}\\
{\normalsize
$^1$North China Electric Power University}\\
{\normalsize
$^2$University of Science and Technology of China\,\,\,$^3$Stanford University}\\
}

\begin{document}

\maketitle

\fancyhead{}
\renewcommand{\headrulewidth}{0pt}

\begin{abstract}
One-dimensional (1D) variable-length visual tokenizers enable adaptive compression by varying the number of tokens, allowing downstream autoregressive (AR) models to flexibly trade off generation quality against computational cost using a single tokenizer. However, existing approaches based on nested dropout often fail to fully exploit the representational capacity of the tokenizer, resulting in suboptimal performance in both image reconstruction and generation. In this work, we introduce NesTok, a nested self-alignment framework tailored to dynamic visual tokenizers. NesTok introduces cross-length training, which jointly optimizes reconstruction across token lengths while using the full-length sequence to guide shorter counterparts, enabling shorter token sequences to approach the reconstruction quality of full-length sequences. On ImageNet, NesTok improves substantially over standard training and achieves an rFID score of 0.98. On downstream image generation, it achieves the state-of-the-art gFID score of \textbf{1.46} on ImageNet 256$\times$256 among existing variable-length autoregressive image generation methods. Code
will be available at \url{https://github.com/Jiawei804/NesTok}.
\end{abstract}

\input{sections/1_introduction}

\input{sections/2_related_work}
\input{sections/3_method}

\input{sections/4_experments}

\input{sections/5_conclution}

\bibliography{iclr2027_conference}
\bibliographystyle{iclr2027_conference}

\appendix
\input{sections/Appendix}

\end{document}

%% file: math_commands.tex
\usepackage{amsmath,amsfonts,bm}

\def\eqref#1{equation~\ref{#1}}

\def\1{\bm{1}}

\DeclareMathAlphabet{\mathsfit}{\encodingdefault}{\sfdefault}{m}{sl}
\SetMathAlphabet{\mathsfit}{bold}{\encodingdefault}{\sfdefault}{bx}{n}

%% file: sections/1_introduction.tex
\section{Introduction}

Image generation has achieved remarkable success under both diffusion-based and autoregressive (AR) paradigms~\citep{ho2020denoising, autoprog, lee2022autoregressive, sun2024autoregressive, Lu_2026_CVPR}, with visual tokenizers playing a pivotal role in this progress. Recent studies~\citep{titok, he2025rear, tatitok, xiong2025gigatok} have explored one-dimensional (1D) visual tokenization to achieve higher compression ratios while preserving reconstruction fidelity. This formulation is conceptually closer to tokenization in natural language and has attracted increasing attention. However, most existing 1D tokenizers operate at a fixed token length and cannot adapt compression ratios within a single model. Supporting different computational budgets and target quality levels therefore requires multiple model variants.

Dynamic visual tokenizers address this limitation by mapping images to variable-length latent sequences of continuous embeddings or discrete codes~\citep{one-d-piece, flextok, semanticist, D_ar, huang2025spectralar, alit, wang2025visual}, allowing the token budget to be adjusted according to computational constraints or image complexity. A prevalent strategy is nested dropout~\citep{flextok}, which randomly truncates the latent sequence and remains the prefix during training, encouraging an ordered representation in which early tokens capture the most important visual information. Representative approaches include FlexTok~\citep{flextok}, Semanticist~\citep{semanticist} and One-D-Piece~\citep{one-d-piece}. When shorter representations correspond to prefixes of the full-length sequence, a downstream AR generator can be trained exclusively on full-length sequences and generate prefixes of different lengths at inference time, enabling flexible trade-offs between generation quality and computational cost.

However, flexibility in token length is not necessarily equivalent to effective use of additional tokens. Prior studies~\citep{retok, he2025rear, flextok, semanticist} have shown that increasing sequence length may yield diminishing improvements in downstream AR generation quality or even degrade it. This observation suggests \textit{inefficient information allocation} across token lengths: shorter prefixes may provide insufficient representations, while later tokens may fail to contribute useful complementary information. Effective variable-length tokenization therefore requires both informative prefixes and progressive refinement through additional tokens. To improve information allocation, ReTok~\citep{retok} introduces redundant token padding, whereas CaTok~\citep{catok} couples token selection intervals with time intervals in the MeanFlow objective to encourage a more balanced distribution of information.

\begin{figure*}[t!]
    \vspace{-5pt}
    \centering
    \includegraphics[width=.99\linewidth]{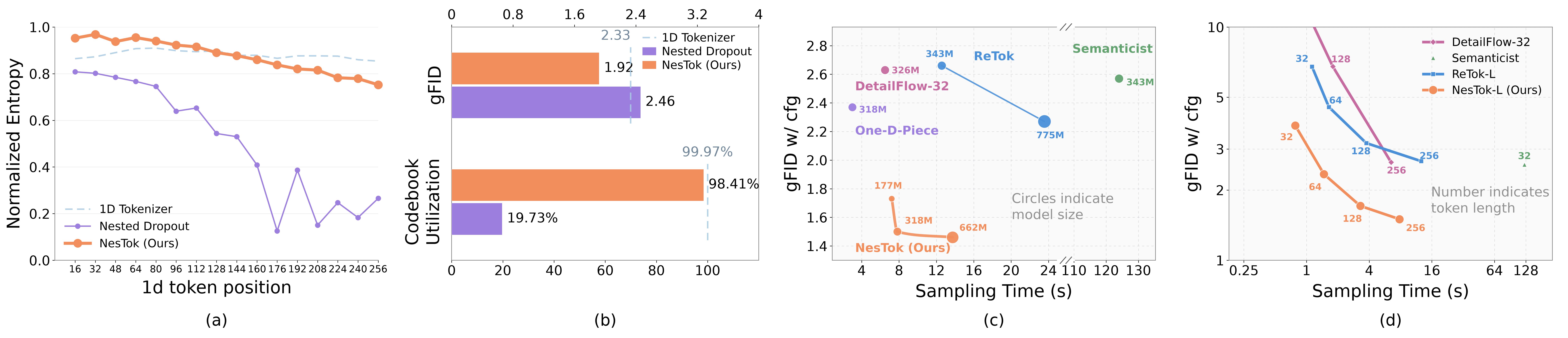}
    \caption{\textbf{Codebook statistics and AR generation performance on ImageNet-1K.} (a): Normalized entropy of the code distribution at each token position, computed over the training dataset and averaged within groups of 16 positions. (b): Average codebook utilization across token positions and gFID. (c): Sampling efficiency with different methods. (d): Sampling efficiency for different token lengths with different methods.
    }
    \label{fig:codebook_distribution}
    \vspace{-10pt}
\end{figure*}

We analyze codebook statistics and AR generation performance on ImageNet-1K, as shown in Fig.~\ref{fig:codebook_distribution}. Fig.~\ref{fig:codebook_distribution} (a) shows that training with nested dropout results in substantially reduced code diversity at later token positions, reflected in their low normalized entropy. The result suggests underutilization of later tokens, potentially limiting their ability to complement the early tokens. Fig.~\ref{fig:codebook_distribution} (b) further shows that nested dropout achieves a gFID of only 2.46, suggesting that concentrated code distribution and tail-token collapse lead to poor generation quality.

Beyond information allocation, decoder design also affects the simplicity and efficiency of variable-length image generation. Recent work has explored diffusion or flow-based decoders~\citep{flextok, semanticist, D_ar} to improve generation quality, potentially introducing additional architectural complexity and decoding latency. We focus on discrete variable-length tokenizers with ViT decoders, retaining the standard next-token prediction paradigm for downstream autoregressive generation. Within this setting, a central challenge is to learn representations that remain effective across token lengths: shorter prefixes should capture global visual content, while later tokens should contribute details for progressive refinement. Meeting these requirements calls for training strategies that coordinate learning across token lengths while supporting downstream AR modeling. This motivates us to revisit the representations learned by variable-length tokenizers: \textit{\textbf{How can a single tokenizer learn informative prefixes and useful refinements that together support effective autoregressive generation?}}

In this work, we introduce NesTok, a variable-length 1D tokenizer built on ViT and trained through \textit{nested self-alignment learning}, which substantially improves autoregressive image generation. We first establish a training recipe for learning a variable-length tokenizer from scratch, without relying on any external distillation. Building on this recipe, we propose a \textit{nested self-aligned tokenizer} that jointly optimizes reconstruction across token lengths while aligning the semantic representations of shorter sequences with those of their full-length counterparts. At each training iteration, representations associated with shorter token sequences are encouraged to align with the full-length sequence in latent space. We empirically demonstrate that this mechanism alleviates the \textit{collapse of tail-token information} associated with nested dropout. Fig.~\ref{fig:codebook_distribution} (a) shows that NesTok preserves code diversity across token positions, with a gradual entropy decline consistent with progressive refinement. In Fig.~\ref{fig:codebook_distribution} (b), NesTok increases average codebook utilization from 19.73\% to 98.41\% and reduces gFID from 2.46 to 1.92, outperforming the fixed-length baseline (2.33). Fig.~\ref{fig:codebook_distribution} (c) demonstrates a favorable trade-off between quality and latency across model sizes. With a single AR generator, NesTok also consistently improves generation quality as token length increases, achieving lower gFID at comparable sampling times (Fig.~\ref{fig:codebook_distribution} (d)).

We validate the effectiveness of NesTok on ImageNet-1K 256$\times$256 image generation. NesTok achieves a reconstruction FID (rFID) of \textbf{0.98} with 256 latent tokens (vs. 1.93 for our variable-length 1D baseline). When paired with LlamaGen, NesTok achieves a generation FID (gFID) of \textbf{1.46} with vanilla AR models, without introducing additional diffusion components. Furthermore, generation quality improves consistently as the number of tokens increases across token lengths. We hope that this work encourages further exploration of ViT-based variable-length visual tokenizers for autoregressive image generation. Our contributions are as follows:

\vspace{-0.05in}
\begin{itemize}[leftmargin=*,itemsep=0mm]
    \item We introduce \textbf{NesTok}, a variable-length 1D ViT tokenizer trained without external distillation to support flexible reconstruction and effective autoregressive generation.

    \item We propose \textbf{cross-length joint training} and \textbf{nested self-alignment} to improve token utilization and establish a coarse-to-fine ordering of visual information.

    \item NesTok achieves an rFID of \textbf{0.98} with 256 tokens and a gFID of \textbf{1.46} on ImageNet-1K $256\times256$ using vanilla AR generation. A trained generator supports different token lengths, with generation quality improving consistently as more tokens are generated.

\end{itemize}

%% file: sections/2_related_work.tex
\section{Related Work}
\noindent\textbf{Image Tokenizers. }Image tokenizers compress high-dimensional images into compact continuous or discrete latent representations for efficient generative modeling. Variational autoencoders (VAEs)~\citep{vae,mactok,lightingdit} map images into continuous latent spaces and are optimized using a reconstruction loss with KL-divergence regularization. VQ-VAE~\citep{vq-vae,vq-vae2} and VQGAN~\citep{vq-gan} instead encode images into fixed 2D grids of discrete tokens. Subsequent studies~\citep{magvit,magvit-v2,fsq,rq-vae,ibq} introduce advanced quantization techniques to improve reconstruction fidelity, codebook utilization, and downstream generation performance. Recently, 1D tokenizers~\citep{titok,softvq,semtok} have transformed images into compact token sequences, substantially reducing the spatial redundancy inherent in conventional 2D grids. The latest work, EOSTok~\citep{chu2026end} unifies the conventional two-stage tokenizer and generator pipeline through end-to-end training.

\noindent\textbf{1D Variable-length Visual tokenizers. }Compared with fixed-length 1D tokenizers, flexible visual tokenizers aim to encode images into variable-length token sequences using a single model~\citep{flextok, one-d-piece, huang2025spectralar}. Existing approaches can be categorized by their decoding mechanisms: \textbf{1) ViT decoders}: One-D-Piece~\citep{one-d-piece} employs nested tail dropping together with two-stage training and external distillation. SpectralAR~\citep{huang2025spectralar} introduces a spectral information loss to encourage causal structure in 1D token sequences. ReTok~\citep{retok} improves dynamic tokenization through redundant tokens and hierarchical semantic regularization. \textbf{2) Diffusion/Flow Decoders}: FlexTok~\citep{flextok} replaces the ViT decoder with a rectified-flow decoder. FlexTok achieves the best gFID with 32 tokens, while generating additional tokens degrades generation quality. D-AR~\citep{D_ar} and CaTok~\citep{catok} associate time steps in the decoding process with intervals of the token sequence to enable efficient image reconstruction. 
Despite these advances, unbalanced information allocation across token lengths remains a challenge, limiting the benefits of variable-length tokenization for downstream autoregressive generation. We propose a nested self-aligned training framework, which requires no external distillation and aims to alleviate this imbalance while improving ordered representations suitable for autoregressive modeling.

\noindent\textbf{Modern Autoregressive Visual Generation. }Inspired by GPT-style language models, early autoregressive image generators built on discrete visual tokenizers, such as VQ-VAE and VQGAN~\citep{llamagen,vq-vae,vq-gan}, flatten two-dimensional token grids into one-dimensional sequences and generate tokens in raster-scan order using causal attention and next-token prediction (NTP). However, this fixed ordering restricts the use of bidirectional spatial context. MAR~\citep{mar} and MaskGIT~\citep{magvit} employ masked prediction with bidirectional attention to incorporate context from visible tokens during iterative generation. VAR~\citep{var} adopts next-scale prediction, reformulating image generation as a coarse-to-fine process that progressively introduces finer visual details. RAR~\citep{rar} and RandAR~\citep{randar} further explore randomized token prediction orders to improve contextual modeling. Although effective, these approaches introduce changes to the generation order, attention pattern, or prediction target of vanilla autoregressive modeling. Unlike these approaches that adapt the generator to visual data, we focus on reshaping the visual representations themselves. Our method encourages more balanced information allocation across token lengths while improving the compatibility of visual tokens with a vanilla autoregressive model without requiring additional changes to the generation mechanism.

%% file: sections/3_method.tex
\section{Method}
1D tokenizers have attracted considerable attention for their flexible compression capabilities. However, enabling efficient downstream autoregressive modeling with variable-length tokenizers remains challenging. Our goal is to learn variable-length token sequences in which short prefixes provide effective image representations and additional tokens contribute complementary refinements. To this end, NesTok combines cross-length joint training with nested self-alignment. These mechanisms aim to improve information utilization and encourage a coarse-to-fine ordering that enhances the \textit{autoregressive compatibility} of the learned tokens. We first introduce the architecture of our ViT variable-length 1D tokenizer (Section~\ref{sec:1D_Vision_Transformer_Tokenizer}), then present our nested self-alignment training method (Section~\ref{sec:Nested_Self-Alignment_Learning}), and finally introduce variable-length autoregressive modeling (Section~\ref{sec:Dynamic_Autoregressive_Model}).

\subsection{1D Vision Tokenizer and Autoregressive Model}
\label{sec:1D_Vision_Transformer_Tokenizer}

Our 1D variable-length tokenizer is based on One-D-Piece~\citep{one-d-piece}, an input image $\mathbf{x}\in\mathbb{R}^{H\times W\times C}$ is partitioned into non-overlapping patches of size $f\times f$ and mapped to a sequence of patch embeddings $\mathbf{x}_{\mathrm{p}}\in\mathbb{R}^{N\times D}$, where $N=HW/f^2$, $C$ is the number of image channels, and $f=16$. These embeddings are concatenated with latent tokens $\mathbf{q}\in\mathbb{R}^{K\times D}$ and processed by the encoder $\mathcal{E}_{\psi}$ to obtain $[\mathbf{h}_{\mathcal{E}},\mathbf{z}]=\mathcal{E}_{\psi}([\mathbf{x}_{\mathrm{p}},\mathbf{q}])$. We discard the encoded patch representations $\mathbf{h}_{\mathcal{E}}$ and retain the 1D latent representations $\mathbf{z}$, which are subsequently quantized as $\mathbf{z}_{q}=\mathcal{Q}(\mathbf{z})$ by looking up the closest entry using a vector quantizer. The quantized latent tokens are then concatenated with mask tokens $\mathbf{m}_{\mathrm{p}}\in\mathbb{R}^{N\times D}$ and passed to the decoder $\mathcal{D}_{\phi}$ for reconstruction, yielding $[\varnothing,\mathbf{x}_{r}]=\mathcal{D}_{\phi}([\mathbf{z}_{q},\mathbf{m}_{\mathrm{p}}])$. Here, $\mathbf{m}_{\mathrm{p}}$ is formed by repeating a mask token $N$ times, and $\varnothing$ denotes the discarded latent tokens.

Consider a variable-length 1D tokenizer that encodes an image
into a quantized token sequence
$\mathbf{z}_{q}=[\mathbf{z}_{1},\mathbf{z}_{2},\ldots,\mathbf{z}_{K}]$
using an encoder $\mathcal{E}$ and a quantizer $\mathcal{Q}$.
Here, $K$ denotes the full sequence length, and $\mathbf{z}_{k}$
is the token at position $k$.
During training, we uniformly sample an integer
$k\in\{1,\ldots,K\}$ and truncate the sequence to retain
a prefix of length $k$, denoted by
$\mathbf{z}_{q}^{(k)}=[\mathbf{z}_{1},\ldots,\mathbf{z}_{k}]$.
The retained prefix is then passed to the decoder
$\mathcal{D}$ to reconstruct the input image.
For the discrete tokenizer considered here, we use a composite
training objective comprising four terms~\citep{yu2024image}:
\begin{equation}
\mathcal{L}_{\mathrm{rec}}
=
\mathcal{L}_{\mathrm{mse}}
+
\mathcal{L}_{\mathrm{perc}}
+
\mathcal{L}_{\mathrm{quant}}
+
\mathcal{L}_{\mathrm{adv}}.
\label{eq:loss_standard}
\end{equation}
Here, $\mathcal{L}_{\mathrm{mse}}$ measures the mean squared error between the reconstructed image $\hat{\mathbf{x}}$ and the input image $\mathbf{x}$. The perceptual loss $\mathcal{L}_{\mathrm{perc}}$ combines LPIPS with a feature perceptual loss computed using ConvNeXt-S. The quantization loss $\mathcal{L}_{\mathrm{quant}}$ comprises the codebook and commitment terms, while $\mathcal{L}_{\mathrm{adv}}$ is an adversarial loss that encourages visually realistic reconstructions. In addition to gradient-based updates driven by the codebook loss, we apply auxiliary usage-adaptive codebook updates guided by exponential moving average (EMA) estimates of codebook usage~\citep{zheng2023online}.

Although most discrete visual tokenizers are trained with this form~\cite{alitok}, their performance is highly sensitive to the training recipe. We therefore establish a standard training recipe as our baseline, with details provided in Section~\ref{sec:experiments}.

\textbf{Autoregressive Compatibility. }The inherent spatial continuity of images introduces redundancy across neighboring patches. Although 1D ViT tokenizers reorganize spatial features into a latent sequence, their encoders jointly optimize token representations for reconstruction without explicitly encouraging an information ordering suitable for autoregressive prediction. Consequently, strong reconstruction performance does not necessarily imply strong \textit{autoregressive compatibility}. Nested dropout encourages reconstruction from shorter prefixes, but later tokens are supplied to the decoder less frequently, potentially contributing to unbalanced information allocation and limiting the benefits of larger token budgets. Therefore, we propose a nested self-alignment training framework to jointly improve information allocation and token ordering (Section~\ref{sec:Nested_Self-Alignment_Learning}).

\begin{figure*}[t!]
    \vspace{-15pt}
    \centering
    \includegraphics[width=.99\linewidth]{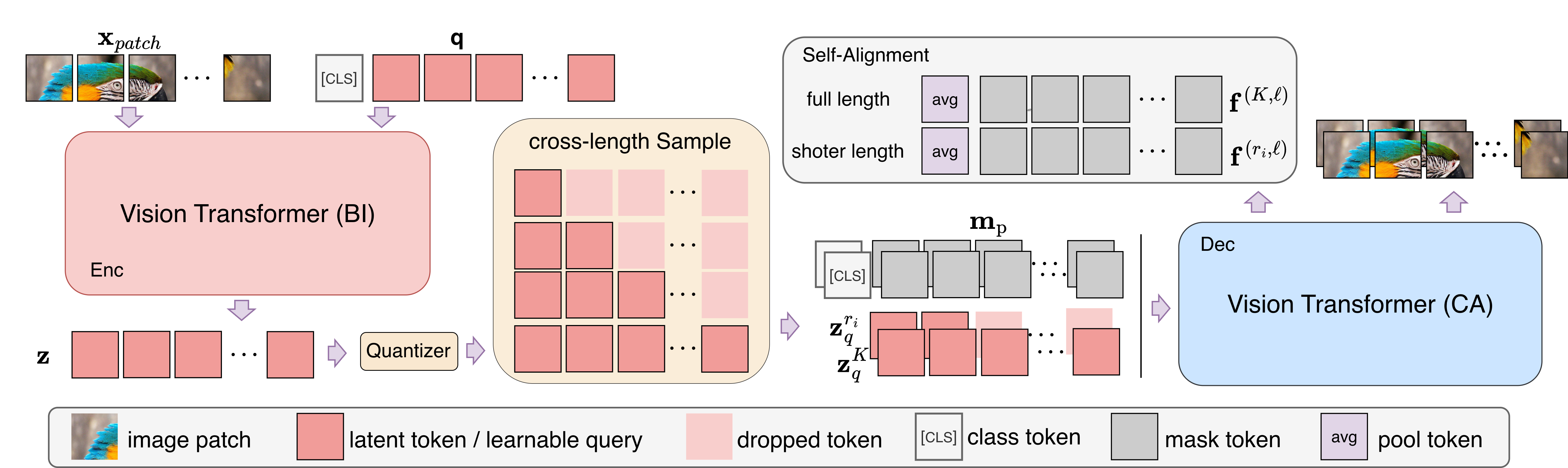}
    \caption{\textbf{Overview of the nested self-aligned training pipeline.} We train a 1D variable-length tokenizer using two components: (1) cross-length sampling, which jointly optimizes reconstruction from a full-length sequence and a sampled shorter sequence at each iteration; and (2) a nested self-alignment loss, which aligns latent representations across the two lengths to promote cross-length consistency and improve reconstruction.
}
    \label{fig:method}
    \vspace{-10pt}
\end{figure*}

\subsection{Nested Self-Alignment Learning}
\label{sec:Nested_Self-Alignment_Learning}

\textbf{Cross-Length Joint Training. }We introduce a cross-length joint training strategy to improve training across token lengths. For an input image $\mathbf{x}$, let $\mathbf{z}_{q}^{(K)}=[\mathbf{z}_{1},\ldots,\mathbf{z}_{K}]$ denote full-length quantized latent sequence. At each training iteration $i$, we retain the full-length sequence $\mathbf{z}_{q}^{(K)}$ and sample a shorter length sequence $\mathbf{z}_{q}^{(r_i)}$, where $r_i\sim \{1,\ldots,K\}$. The resulting sequence $\mathbf{z}_{q}^{(r_i)}$ is therefore a prefix of $\mathbf{z}_{q}^{(K)}$. Both sequences are decoded, and their losses are accumulated to jointly optimize the tokenizer:
\begin{equation}
\mathcal{L}_{\mathrm{joint}}
=
\mathcal{L}_{\mathrm{rec}}(\mathbf{x};K)
+
\mathcal{L}_{\mathrm{rec}}(\mathbf{x};r_i),
\label{eq:multi_sample}
\end{equation}

The two samples share all tokenizer parameters. The full-length sequence ensures that every latent token position is retained at least one decoding pass per iteration, while the shorter sequence trains the tokenizer to reconstruct images with a variable length to compress the reconstruction information.

\textbf{Nested Self-Alignment Loss. }Although cross-length joint training provides consistent supervision for full length reconstruction, it does not explicitly encourage feature consistency across token lengths. Empirically, we observe that shorter prefixes improve slowly and retain a substantial reconstruction gap relative to full-length sequences. Despite parameter sharing, improvements in full length reconstruction do not necessarily translate into comparable improvements at shorter lengths. This motivates an explicit mechanism for transferring alignment information at feature level across lengths. To this end, we introduce a nested self-alignment loss. With $n\in\{r_i,K\}$, we extract the output feature of the $\ell$-th layer of the decoder:
\begin{equation}
\mathbf{h}_{z}^{(n,\ell)},
\mathbf{h}_{\mathrm{p}}^{(n,\ell)}
=
\mathcal{D}_{\phi}^{(1:\ell)}
(
\mathbf{z}_{q}^{(n)},
\mathbf{m}_{\mathrm{p}}
),
\label{eq:alignment_decoder_features}
\end{equation}
where $\mathcal{D}_{\phi}^{(1:\ell)}$ denotes Parameters of the first $\ell$ layers of the decoder and $\mathbf{h}_{z}^{(n,\ell)}\in\mathbb{R}^{n\times d}$ and $\mathbf{h}_{\mathbf{p}}^{(n,\ell)}\in\mathbb{R}^{N\times d}$ denote the features of latent tokens and mask tokens, respectively. Unless otherwise specified, we apply alignment at the first decoder layer, i.e., $\ell=1$.

Since the two sampled sequences contain different numbers of latent tokens, we apply average pooling along the token dimension to obtain a pooled latent feature, $\bar{\mathbf{h}}_{z}^{(n,\ell)}=\operatorname{Avg}(\mathbf{h}_{z}^{(n,\ell)})$, where $\mathbf{h}_{z}^{(n,\ell)}$ contains the $n$ latent-token features. We then concatenate this feature, treated as a single token, with the mask-token features along the token dimension to form $\mathbf{f}^{(n,\ell)}=\operatorname{Concat}(\bar{\mathbf{h}}_{z}^{(n,\ell)},\mathbf{h}_{\mathrm{p}}^{(n,\ell)})\in\mathbb{R}^{(N+1)\times d}$. We define the nested self-alignment loss as:
\begin{equation}
\mathcal{L}_{\mathrm{align}}
=
1-
\frac{1}{N+1}
\sum_{j=1}^{N+1}
\operatorname{sim}(
\mathbf{f}_{j}^{(r_i,\ell)},
\operatorname{stopgrad}[
\mathbf{f}_{j}^{(K,\ell)}
]),
\label{eq:nested_self_alignment}
\end{equation}
The full-length features serve as an online alignment target for shorter sequences. This supervision in feature space encourages shorter sequences to rapidly align their representations with those of the full length sequences, improving reconstruction quality at shorter token length without requiring an additional teacher network.

\textbf{Overall objective of Nested Self-Aligned Learning. }As shown in Fig.~\ref{fig:method}, our training framework combines cross-length joint training with nested self-alignment. Cross-length joint training decodes the full-length sequence at every iteration, so every position receives reconstruction supervision. Tail tokens can therefore reduce the full-length loss only by adding details that the prefix lacks, which makes them residual refinements rather than redundant copies. Second, the self-alignment loss pulls the decoder features of each prefix toward those of the full-length sequence, which serves as a stop-gradient target. Since the prefix is a prefix of the full sequence, the most effective way to shrink this gap is to place the information  in the earliest tokens  that best approximates the full representation, such as global layout and semantics, and to leave what the prefix cannot capture to later tokens. The overall training objective is:
\begin{equation}
\mathcal{L}_{\mathrm{total}}
=
\mathcal{L}_{\mathrm{joint}}
+
\mathcal{L}_{\mathrm{align}}
\label{eq:overall_training_objective}
\end{equation}

\subsection{Variable-length Autoregressive Model}
\label{sec:Dynamic_Autoregressive_Model}

Variable-length AR generation relies on an ordered token sequence, in which every prefix is a progressively refined description of the same image. Nested dropout provides only a weak form of this ordering. Because early positions are retained under almost every truncation, they are pushed to carry the most important content. Later positions, however, are rarely supervised. As a result, the sequence tends to become \textit{informative head, idle tail} rather than coarse-to-fine.
NesTok strengthens the ordering in two complementary ways. For the downstream generator, this means that early predictions decide the global content, and later predictions refine it conditioned on that content. This ordering is consistent with next-token prediction.

\textbf{Autoregressive Modeling. }We model discrete visual token sequences through standard autoregressive next-token prediction:
\begin{equation}
p_\theta(\mathbf{z}_{1:n})
=
\prod_{i=1}^{n}
p_\theta(\mathbf{z}_i \mid \mathbf{z}_{<i}),
\end{equation}
where $\theta$ denotes the model parameters and $n$ is the token sequence length. We train the model on full-length sequences by minimizing the cross-entropy loss. At inference time, a desired token budget $n\leq K$ can be specified, where $K$ is the maximum sequence length. The same generator then samples an $n$-token prefix, which is decoded into an image by the tokenizer decoder. Thus, a single trained generator supports variable-length inference, enabling flexible trade-offs between generation quality and computational cost. We additionally employ KV-cache to accelerate autoregressive sampling.

\input{resources/main_table}

%% file: resources/main_table.tex
\begin{table}[t]
\vspace{-10pt}
\centering\small
\setlength{\tabcolsep}{5pt}
\caption{System-level comparison of different tokenizers and generation models on ImageNet 256$\times$256. 
$\downarrow$ and $\uparrow$ indicate whether lower or higher values are better. We categorize tokenizers into three groups: 2D tokenizers, fixed-length 1D tokenizers, and variable-length 1D tokenizers.}
\label{tab:main_results}
\vspace{-5pt}

\resizebox{0.999\textwidth}{!}{%
\begin{tabular}{l|cccc|cc|cc|cc}
\toprule

\multirow{2}{*}{\textbf{Method}}
& \multicolumn{4}{c|}{\textbf{Tokenizer}}
& \multicolumn{2}{c|}{\textbf{Generator}}
& \multicolumn{2}{c|}{\textbf{w/o guidance}}
& \multicolumn{2}{c}{\textbf{w/ guidance}}
\\

\cmidrule(lr){2-5}
\cmidrule(lr){6-7}
\cmidrule(lr){8-9}
\cmidrule(lr){10-11}

& \textbf{Type}
& \textbf{\#Params}
& \textbf{\#Tokens}
& \textbf{rFID}$\downarrow$
& \textbf{Type}
& \textbf{\#Params}
& \textbf{gFID}$\downarrow$
& \textbf{IS}$\uparrow$
& \textbf{gFID}$\downarrow$
& \textbf{IS}$\uparrow$
\\

\midrule
\rowcolor{blue!10}
\multicolumn{11}{c}{\textbf{2D Tokenization}} \\

\midrule

DiT-XL/2~\citep{dit}
& SD-VAE
& 84M
& 256
& 0.62
& Diff.
& 675M
& 9.62
& 121.5
& 2.27
& 278.2     
\\

REPA-XL/2~\citep{repa}
& KL
& 84M
& 1024
& 0.62
& Diff.
& 675M
& 5.90
& 157.8
& 1.42
& 305.7
\\

Lightning-DiT-XL~\citep{lightingdit}
& KL
& 84M
& 1024
& 0.28
& Diff.
& 675M
& 2.17
& 205.6
& 1.35
& 295.3
\\

MAR-L~\citep{mar}
& KL
& 66M
& 256
& 0.87
& MAR Diff.
& 479M
& 2.60
& 221.4
& 1.78
& 296.0
\\

VQGAN~\citep{vq-gan}
& VQ
& 23M
& 256
& 4.98
& AR
& 1.4B
& 15.78
& 74.3
& -
& -
\\

MaskGIT~\citep{chang2022maskgit}
& VQ
& 66M
& 256
& 2.28
& Mask
& 227M
& 6.18
& 182.1
& -
& -
\\

LlamaGen-XL~\citep{llamagen}
& VQ
& 72M
& 256
& 0.94
& AR
& 775M
& 14.77
& 80.8
& 2.62
& 244.1
\\

RAR-L~\citep{rar}
& VQ
& 66M
& 256
& 2.28
& AR
& 461M
& 5.39
& 149.1
& 1.70
& 299.5
\\

IBQ-L~\citep{ibq}
& IBQ
& 128M
& 256
& 1.37
& AR
& 649M
& -
& -
& 2.45
& 267.5
\\

VAR-d20~\citep{var}
& MSRQ
& 109M
& 680
& 0.90
& VAR
& 600M
& -
& -
& 2.57
& 302.6
\\

AliTok-L~\citep{alitok}
& VQ
& 390M
& 273
& 0.86
& AR
& 318M
& 1.98
& 200.8
& 1.38
& 326.2
\\

\midrule
\rowcolor{blue!10}
\multicolumn{11}{c}{\textbf{1D Tokenization}} \\

\midrule

TiTok-L~\citep{titok}
& VQ
& 641M
& 32
& 2.21
& Mask
& 177M
& 3.15
& 173.0
& 2.77
& 199.8
\\

GigaTok~\citep{xiong2025gigatok}
& VQ
& 622M
& 256
& 0.81
& AR
& 111M
& -
& -
& 3.26
& 221.0
\\

SoftVQ-VAE-L~\citep{softvq}
& KL
& 176M
& 64
& 0.61
& Diff.
& 675M
& 5.83
& 141.3
& 2.93
& 268.5
\\

ResTok~\citep{restok}
& VQ
& 662M
& 128
& 1.28
& HAR
& 326M
& -
& -
& 2.34
& 257.8
\\

MacTok~\citep{mactok}
& KL
& 675M
& 128
& 0.43
& Diff.
& 326M
& 3.12
& 186.2
& 1.50
& 299.8
\\

SemTok-XL~\citep{semtok}
& VQ
& 2.35B
& 256
& 0.88
& Mask
& 746M
& -
& -
& 2.54
& 305.6
\\

SemTok-XXL
& VQ
& 2.35B
& 256
& 0.88
& Mask
& 1.2B
& -
& -
& 2.34
& 310.5
\\

SpectralAR-d20~\citep{huang2025spectralar}
& VQ
& -
& 64
& 4.03
& AR
& 600M
& -
& -
& 2.49
& 305.4

\\

D-AR-L~\citep{D_ar}
& VQ
& 300M
& 256
& 1.52
& AR
& 343M
& -
& -
& 2.44
& 262.9

\\

D-AR-XL
& VQ
& 300M
& 256
& 1.52
& AR
& 775M
& -
& -
& 2.09
& 298.4

\\

EOSTok-H~\citep{chu2026end}
& IBQ
& 388M
& 256
& 0.71
& AR
& 644M
& {1.48}
& {239.5}
& {1.38}
& 265.7
\\

\midrule
\rowcolor{blue!10}
\multicolumn{11}{c}{\textbf{1D Variable-length Tokenization}} \\

\midrule

FlexTok d18-d18-32~\citep{flextok}
& FSQ
& 950M
& 1-256
& 1.61
& AR
& 1.33B
& -
& -
& 2.02
& -
\\

FlexTok d18-d28-32
& FSQ
& 2.5B
& 1-256
& 1.45
& AR
& 1.33B
& -
& -
& 1.86
& -
\\

Semanticist-XL-32~\citep{semanticist}
& KL
& -
& 1-256
& \textbf{0.78}
& AR Diff.
& 343M
& -
& -
& 2.57
& 254.0
\\

One-D-Piece~\citep{one-d-piece}
& VQ
& 641M
& 1-256
& 1.08
& Mask
& 318M
& -
& -
& 2.35
& 224.4
\\

DetailFlow-32~\citep{detailflow}
& VQ
& 64M
& 1-256
& 0.80
& AR
& 326M
& -
& -
& 2.75
& 250.8
\\

ReTok~\citep{retok}
& VQ
& 232M
& 1-256
& 1.01
& AR
& 775M
& -
& -
& 2.27
& 245.9
\\

\midrule

\textbf{NesTok-B}
& VQ
& 390M
& 1-256
& 0.98
& AR
& 177M
& 2.59
& 181.6
& 1.73
& 250.5
\\

\textbf{NesTok-L}
& VQ
& 390M
& 1-256
& 0.98
& AR
& 318M
& 2.08
& 205.6
& 1.50
& 278.9
\\

\textbf{NesTok-XL}
& VQ
& 390M
& 1-256
& 0.98
& AR
& 662M
& \textbf{1.87}
& \textbf{238.9}
& \textbf{1.46}
& \textbf{295.9}
\\

\bottomrule
\end{tabular}
}
\end{table}

%% file: sections/4_experments.tex
\vspace{-5pt}
\section{Experiment}
\label{sec:experiments}
\vspace{-5pt}

\subsection{Implementation Details}\label{sec:training}

\textbf{Model Configuration and Evaluation}
Our 1D variable-length tokenizer is based on One-D-Piece~\citep{one-d-piece} and uses ViT-B as the encoder and ViT-L as the decoder, with a codebook of 4,096 entries and a latent dimension of 32. Our autoregressive model adopts the decoder-only Transformer architecture of LlamaGen~\citep{llamagen}. To accommodate 1D latent sequences, we replace 2D rotary positional embeddings (RoPE) with 1D RoPE~\citep{su2024roformer}. We named our tokenizer as NesTok and denote the corresponding autoregressive models at different scales as NesTok-B/L/XL. Following prior work, we evaluate performance using Fr\'echet Inception Distance (FID), including reconstruction FID (rFID) and generation FID (gFID), as well as Inception Score (IS). We compute gFID on 50,000 generated images using the ADM’s TensorFlow evaluation suite~\citep{dhariwal2021diffusion} and evaluate rFID using the MAR evaluation code~\citep{mar}.

\textbf{Training of Tokenizer. }We train NesTok in three stages without any external teacher for distillation or representation alignment. The nested self-alignment loss is applied throughout training, with a batch size of 256. \textbf{In the first stage}, we jointly optimize the reconstruction, perceptual, and quantization losses for 200K iterations, warming up the learning rate to 1e-4. \textbf{In the second stage}, we disable the LPIPS of the perceptual loss, retaining only the loss computed using ConvNeXt-S. This stage lasts 200K iterations, during which the learning rate decreases from 5e-5 to 2e-5. \textbf{In the third stage}, we additionally introduce the GAN loss to improve the visual fidelity of reconstructions. We train for 150K steps, decreasing the learning rate from 2e-5 to 5e-6.

\textbf{Training of Autoregressive Model. }Following common practice, we train the LlamaGen with a batch size of 2,048 and a learning rate of 4e-4 with a warmup for 100 epochs. We train for 400 epochs totally, corresponding to approximately 250K steps. We apply QK-Norm~\citep{team2024chameleon} in the attention modules and use RMSNorm~\citep{zhang2019root} for normalization. Before AR training, we precompute and cache the discrete visual token sequences using the tokenizer's encoder and quantizer to accelerate training.

\begin{figure*}[t!]
    \vspace{-5pt}
    \centering
    \includegraphics[width=.99\linewidth]{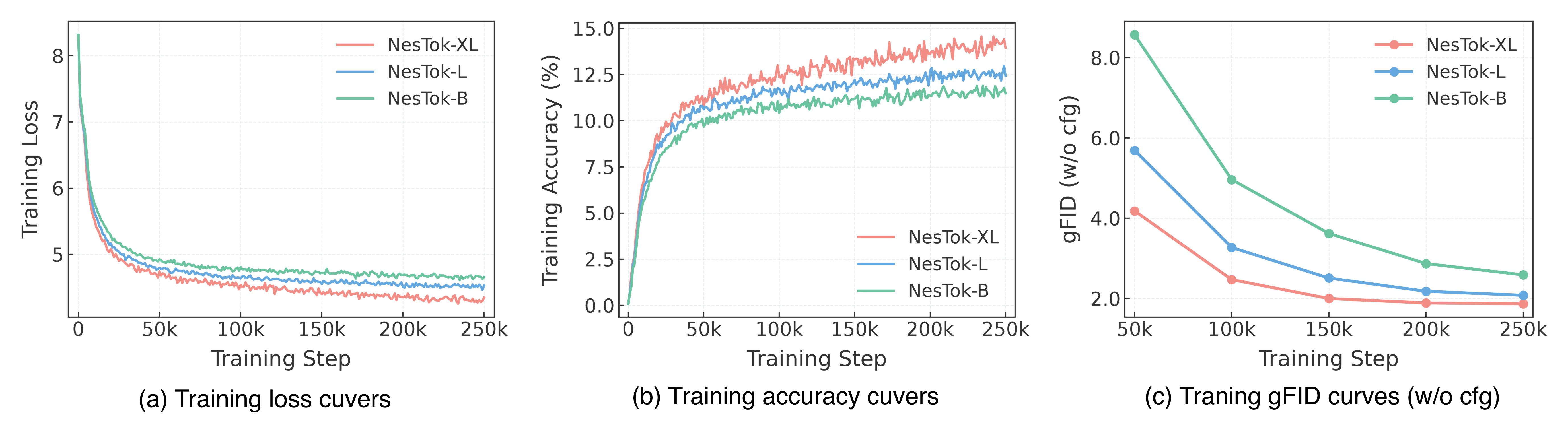}
    \vspace{-5pt}
    \caption{\textbf{Training curves.} (a) Training loss. (b) Training accuracy (\%). (c) gFID without classifier-free guidance across training steps and model sizes.
}
    \label{fig:training_curves}
    \vspace{-15pt}
\end{figure*}

\begin{figure*}[t]
    \vspace{-10pt}
    \centering
    \includegraphics[width=.99\linewidth]{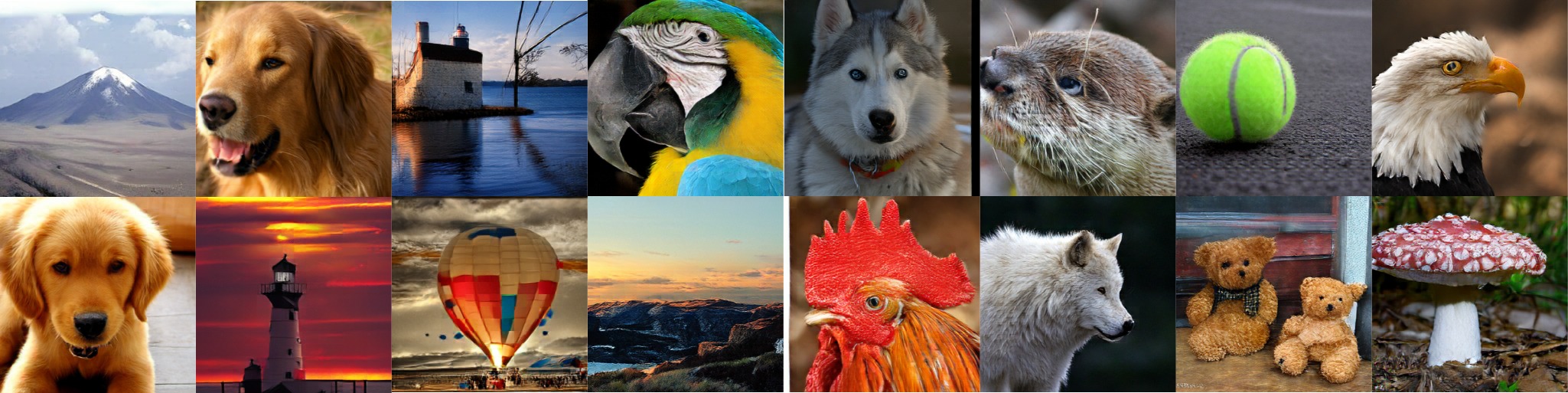}
    \caption{\textbf{Examples of image generation with NesTok-XL on ImageNet $256\times256$. }
}
    \label{fig:visualization_imagenet}
\end{figure*}

\begin{figure}[t!]
    \vspace{-10pt}
    \centering
    \includegraphics[width=.99\linewidth]{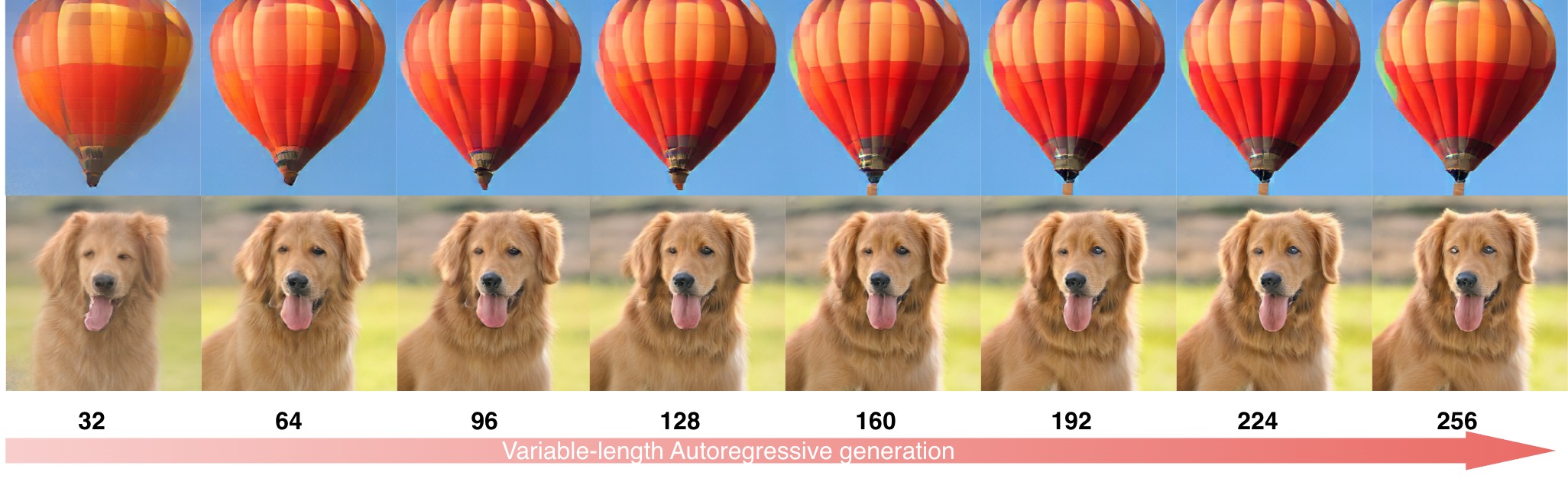}
    \caption{\textbf{Visualization of variable-length generation on ImageNet $256\times256$ resolution.} Images are generated using NesTok-XL with 32-256 tokens.
}
    \label{fig:variable-length_generation}
    \vspace{-5pt}
\end{figure}

\begin{figure}[t!]
    \centering
    \includegraphics[width=.99\linewidth]{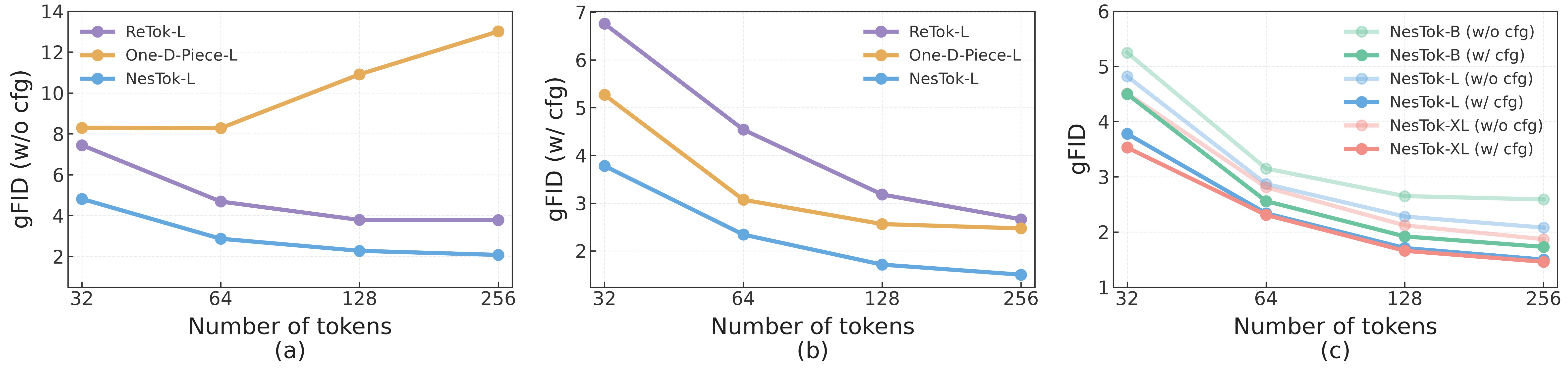}
    \vspace{-5pt}
    \caption{\textbf{gFID w/ and w/o cfg across different methods and model sizes. }
}
    \label{fig:gFID_vs_tokens}
    \vspace{-15pt}
\end{figure}

\subsection{Main Results}

\noindent\textbf{Generation Results on ImageNet 256$\times$256. }
We evaluate NesTok against state-of-the-art methods on ImageNet-1K at $256\times256$ resolution. As shown in Tab.~\ref{tab:main_results}, NesTok achieves strong downstream autoregressive generation performance while maintaining high reconstruction quality. Specifically, NesTok achieves an rFID of 0.98, outperforming One-D-Piece's 1.08 with fewer tokenizer parameters (390M vs.\ 641M) and without external distillation. Among the compared 1D variable-length methods, NesTok-XL achieves the state-of-the-art gFID of \textbf{1.46}, compared with 2.35 for One-D-Piece. It achieves a gFID of \textbf{1.87} without classifier-free guidance, outperforming the other variable-length methods. Although EOSTok achieves the best generation quality among the compared 1D tokenization methods, it jointly trains the tokenizer and AR generator and relies on DINOv2-based distillation. Fig.~\ref{fig:visualization_imagenet} presents qualitative generation results obtained with NesTok. In Fig.~\ref{fig:variable-length_generation}, as the token length increases from 32 to 256, the images exhibit progressively finer details while maintaining consistent global structure.

\noindent\textbf{Variable-length Autoregressive Model. }As shown in Fig.~\ref{fig:gFID_vs_tokens}, for each model configuration, we evaluate a single trained AR generator across token lengths with and without classifier-free guidance (cfg). We compare NesTok with One-D-Piece-L~\citep{one-d-piece} and ReTok-L~\citep{retok}. NesTok-L substantially outperforms both methods across all evaluated token lengths. While One-D-Piece-L exhibits degraded generation quality at larger budgets without cfg, ReTok-L shows diminishing gains: increasing the sequence length from 128 to 256 tokens reduces gFID by only 0.01, from 3.79 to 3.78. Over the same interval, NesTok-L and NesTok-XL reduce gFID by 0.20 and 0.25, respectively. In Fig.~\ref{fig:gFID_vs_tokens} (c), NesTok consistently benefits from increasing the token lengths, with improvements continuing at longer sequence lengths. These results demonstrate the \textit{autoregressive compatibility} of the learned representations and suggest that our nested self-alignment training architecture enables more effective use of additional tokens, including those at later positions.

\textbf{Training Analysis. }
Fig.~\ref{fig:training_curves} illustrates the training dynamics and generation performance of autoregressive models built on NesTok. Fig.~\ref{fig:training_curves} (a) and (b) show steadily decreasing training loss and increasing training accuracy, respectively. Fig.~\ref{fig:training_curves} (c) reports gFID scores using 256 tokens without classifier-free guidance. Scaling up the AR generator consistently improves gFID. These results support the \textit{autoregressive compatibility} of the token sequences learned by NesTok, suggesting that our training strategy produces representations well suited to autoregressive prediction.

\subsection{Ablation Studies. }

\begin{figure*}[t]
    \vspace{-7pt}
    \centering
    \includegraphics[width=.99\linewidth]{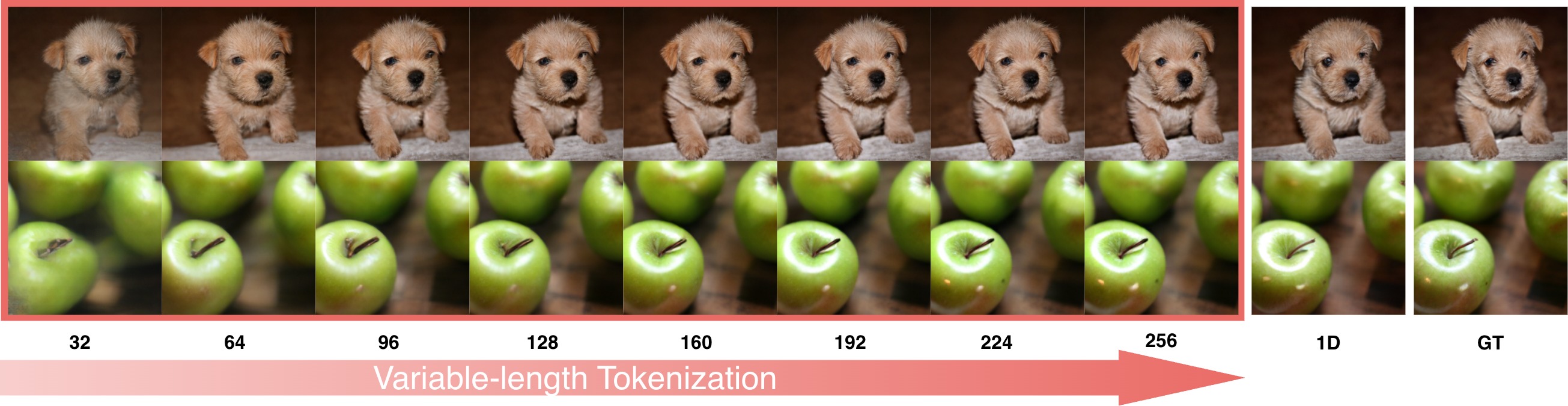}
    \caption{\textbf{Visualization of variable-length reconstruction on ImageNet $256\times256$ resolution.} 1D denotes the 1D Tokenizer with 256 tokens.
}
    \label{fig:variable-length_tokenization}
    \vspace{-5pt}
\end{figure*}

\textbf{Ablation of Components. }
We conduct an ablation study to evaluate the contributions of individual components. As shown in Table~\ref{tab:ablation_training}, the fixed-length 1D Tokenizer achieves the best rFID but does not support variable-length representations. Introducing nested dropout enables variable-length tokenization but degrades reconstruction quality. This variant achieves an AR training accuracy of 46.2\%, yet yields a gFID of 2.46. According to Fig.~\ref{fig:codebook_distribution}, the result suggests that nested dropout is associated with concentrated code usage and tail-token collapse. Cross-length joint training mitigates this issue, ensuring that all latent-token positions participate in reconstruction training. Adding the nested self-alignment loss further promotes \textit{cross-length consistency} and improves reconstruction from shorter sequences. Although this loss yields little improvement in full-length generation quality, it substantially improves generation at shorter lengths. With both components, NesTok achieves a gFID of 1.92, outperforming the 1D Tokenizer's 2.33. Fig.~\ref{fig:variable-length_tokenization} further shows that NesTok preserves the main visual content with short prefixes and restores finer details as the token length increases.

\input{resources/ablation_table}

\begin{wrapfigure}[11]{r}{0.43\textwidth} 
\centering\small
\vspace{-0.15in}
\captionof{table}{
\textbf{Comparison of sampling speed. }Throughput (images/s) measured on a single H800 GPU with a batch size of 128, averaged over ten sampling runs using the official implementation of each method.
}
\vspace{-0.1in}
\resizebox{0.43\textwidth}{!}{
    \begin{tabular}{l|cccc}
    \toprule
    Method & Tokens & Params & gFID & images/s\\
    \midrule
    ReTok-L & 256 & 343M & 2.27 & 10.11\\ 

    One-D-Piece-L & 256 & 318M & 2.35 & 42.67\\
    
    DetailFlow & 256 & 326M & 2.75 & 19.69\\

    Semanticist-L & 32 & 343M & 2.57 & 1.03 \\

    NesTok-L & 256 & 318M & \textbf{1.50} & 16.35\\
    
     \bottomrule
    \end{tabular}
}
\label{tab:sit_imagenet}

\end{wrapfigure}

\textbf{Comparison of sampling speed. }We compare the sampling speed of different methods in terms of throughput. Although One-D-Piece achieves high throughput with its MaskGIT generator, its generation quality is comparatively limited. Semanticist employs a diffusion decoder and achieves a throughput of only 1.03 images per second even with 32 tokens, limiting its suitability for applications requiring rapid responses. NesTok achieves the best gFID among the compared methods while maintaining competitive throughput, offering a favorable trade-off between generation quality and sampling efficiency.

%% file: resources/ablation_table.tex
\begin{table}[t!]
\centering
\small
\setlength{\tabcolsep}{7pt}
\renewcommand{\arraystretch}{1.15}

\caption{\textbf{Ablation study of different training settings.}
We report generation results using LlamaGen-Base (177M) trained for 200 epochs.}
\label{tab:ablation_training}

\resizebox{0.95\textwidth}{!}{%
\begin{tabular}{l|cc|cccc|cccc}
\toprule

\multirow{3}{*}{\textbf{Training Setting}}
& \multicolumn{2}{c|}{\textbf{AR Training}}
& \multicolumn{8}{c}{\textbf{Evaluation}}
\\

\cmidrule(lr){2-3}
\cmidrule(lr){4-11}

& \multirow{2}{*}{\textbf{Loss}$\downarrow$}
& \multirow{2}{*}{\textbf{Acc.}$\uparrow$}
& \multicolumn{4}{c|}{\textbf{rFID}$\downarrow$}
& \multicolumn{4}{c}{\textbf{gFID}$\downarrow$}
\\

\cmidrule(lr){4-7}
\cmidrule(lr){8-11}

& &
& \textbf{32}
& \textbf{64}
& \textbf{128}
& \textbf{256}
& \textbf{32}
& \textbf{64}
& \textbf{128}
& \textbf{256}
\\

\midrule

1D Tokenizer
& 5.37
& 7.9\%
& -
& -
& -
& 0.83
& -
& -
& -
& 2.33
\\

\midrule

+Nested Dropout
& 2.37
& 46.2\%
& 4.28
& 2.88
& 2.07
& 1.93
& 5.98
& 4.64
& 3.12
& 2.46
\\

+Cross-length Sample
& 4.70
& 11.3\%
& 3.70
& 2.14
& 1.43
& 0.98
& 5.76
& 4.16
& 2.36
& 1.93
\\

+Nested Self-Alignment Loss
& 4.68
& 11.4\%
& \textbf{3.49}
& \textbf{2.10}
& \textbf{1.39}
& \textbf{0.98}
& \textbf{5.48}
& \textbf{3.85}
& \textbf{2.18}
& \textbf{1.92}
\\

\bottomrule
\end{tabular}%

}
\vspace{-10pt}
\end{table}

%% file: sections/5_conclution.tex
\section{Conclusion}
We presented NesTok, a variable-length 1D tokenizer trained through nested self-alignment without external distillation. By combining cross-length joint training with feature alignment, our framework mitigates tail-token collapse, increases average codebook utilization to nearly 100\%. The learned representations support coarse-to-fine refinement and exhibit improved autoregressive compatibility. The AR generator trained with NesTok supports multiple token lengths, with generation quality improving consistently as more tokens are generated. The result demonstrates that an effective tokenizer training strategy can improve the variable-length image generation while retaining standard next-token prediction.

%% file: sections/Appendix.tex
\section{Appendix}

\subsection{Training Pseudocode}
The pseudocode for our training procedure is presented below:

\begin{algorithm}[h]
\caption{Nested Self-Alignment Training for NesTok}
\label{alg:nestok}
\textbf{Input}:
Training dataset $\mathcal{X}$; \\
$\mathcal{E}_{\psi}$: encoder;
$\mathcal{D}_{\phi}$: decoder;
$\mathcal{Q}$: vector quantizer; \\
$K$: maximum token length;
$S$: training iterations; \\
$\ell$: alignment layer;
$\mathcal{P}$: sampling distribution over $\{1,\ldots,K\}$.\\
\textbf{Output}: Trained tokenizer
$(\mathcal{E}_{\psi}, \mathcal{Q}, \mathcal{D}_{\phi})$.

\begin{algorithmic}[1]
\STATE Initialize the tokenizer parameters;
\FOR{$i = 1, \ldots, S$}
    \STATE Sample a batch $\mathbf{x}$ from $\mathcal{X}$;
    \STATE Encode and quantize $\mathbf{x}$ to obtain
    the full-length sequence $\mathbf{z}_{q}^{(K)}$;
    \STATE Sample $r_i \sim \mathcal{P}$ and retain the prefix
    $\mathbf{z}_{q}^{(r_i)}$;

    \FOR{$n \in \{K, r_i\}$}
        \STATE Decode $\mathbf{z}_{q}^{(n)}$ to reconstruct
        $\hat{\mathbf{x}}^{(n)}$ and extract features using Eq.~\textcolor{red}{3};
        \STATE Average latent-token features and concatenating with the mask-token features: $\mathbf{f}^{(n,\ell)}$;
        \STATE Compute $\mathcal{L}_{\mathrm{rec}}(\mathbf{x};n)$
        using Eq.~\textcolor{red}{1};
    \ENDFOR

    \STATE Compute the joint reconstruction loss
    $\mathcal{L}_{\mathrm{joint}}$ using ~\textcolor{red}{2};
    \STATE Compute $\mathcal{L}_{\mathrm{align}}$ using Eq.~\textcolor{red}{4},
    with $\mathbf{f}^{(K,\ell)}$ as the stop-gradient target;
    \STATE Compute the total objective
    $\mathcal{L}_{\mathrm{total}}$ using Eq.~\textcolor{red}{5};
\ENDFOR
\end{algorithmic}
\end{algorithm}

\subsection{More Results}

\begin{figure}[h]
    \centering
    \includegraphics[width=.99\linewidth]{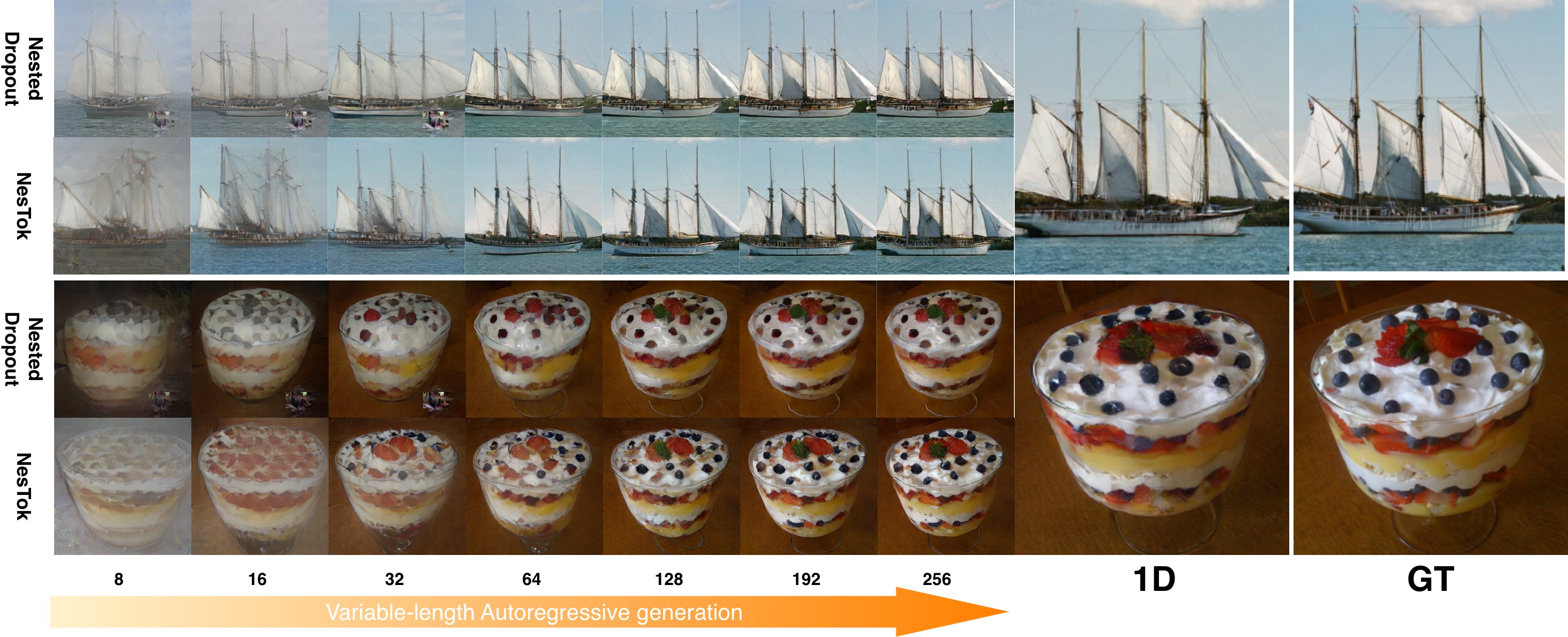}
    \caption{\textbf{Visualization of variable-length generation on ImageNet $256\times256$ resolution.} Images are generated using NesTok-XL with 8-256 tokens and without classifier-free guidance.
}
    \label{fig:variable-length_tokenization1}
    \vspace{-5pt}
\end{figure}

\textbf{Visualization of Variable-length Reconstruction. }
As shown in Fig.~\ref{fig:variable-length_tokenization1}, we compare reconstructions from NesTok and the nested dropout  across token lengths ranging from 8 to 256. Nested dropout exhibits pronounced artifacts with fewer than 32 tokens. Although its reconstruction quality improves as more tokens are retained, a substantial gap in visual fidelity compared to NesTok persists even at longer sequence lengths.

\textbf{Visualization of variable-length Generation}
We visualize images decoded from autoregressively generated token sequences ranging from 8 to 256 tokens. As more tokens are generated, coarse visual content is progressively refined into sharper and more detailed images. For the Granny Smith apple (948), additional tokens refine its shape, surface texture, and lighting. For the bald eagle (22), feather textures become more distinct, while the eyes and beak gain sharper definition. For the promontory (976), later tokens refine the coastline, terrain, and foreground vegetation. These examples illustrate how additional tokens contribute complementary visual details, supporting a coarse-to-fine generation process.
\begin{figure}[h]
    \centering
    \includegraphics[width=.99\linewidth]{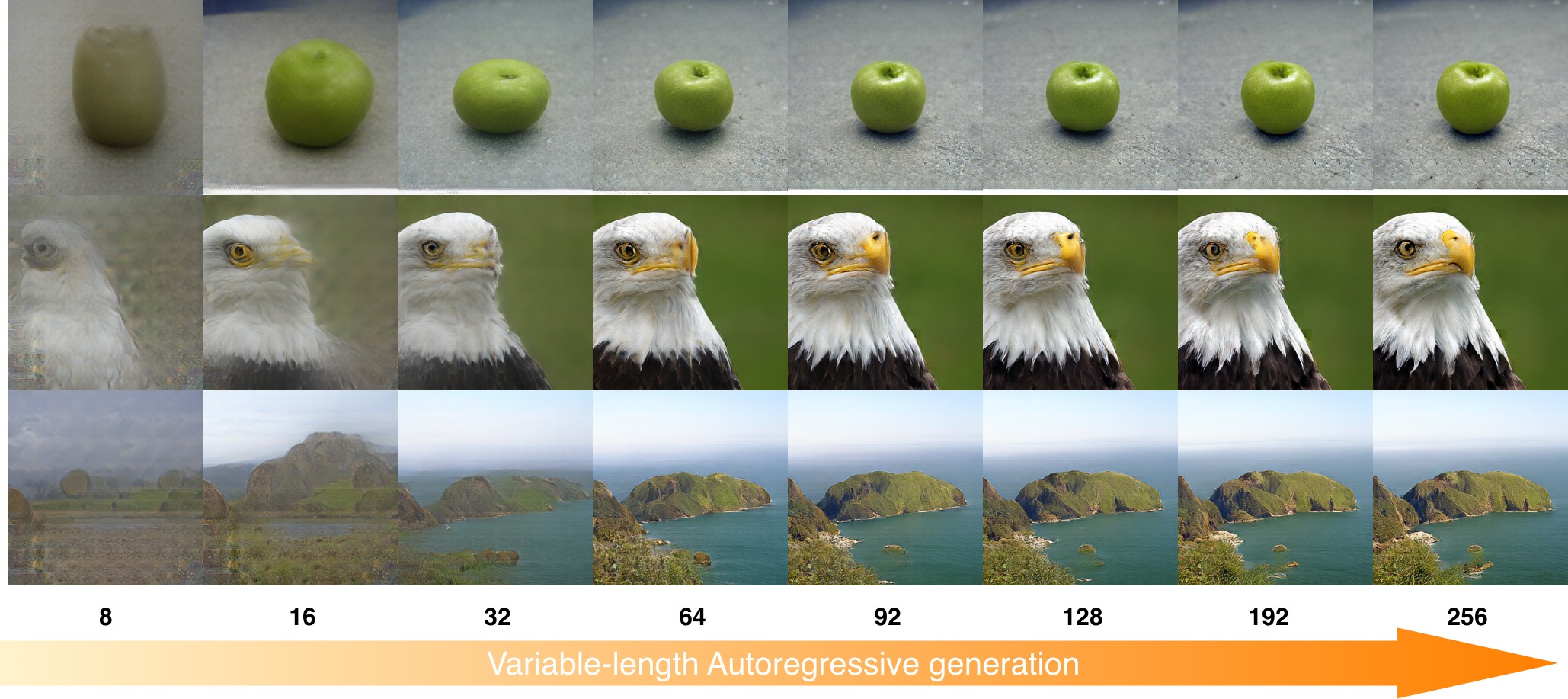}
    \caption{\textbf{Visualization of variable-length generation on ImageNet $256\times256$ resolution.} Images are generated using NesTok-XL with 8-256 tokens and without classifier-free guidance.
}
    \label{fig:variable-length_tokenization}
    \vspace{-5pt}
\end{figure}

\textbf{More reconstruction results. }
Tab.~\ref{tab:rfid_different_tokens} compares reconstruction quality across token lengths on ImageNet at $256\times256$ resolution. NesTok consistently outperforms ReTok-S-B, with rFID improving from 3.49 at 32 tokens to 0.98 at 256 tokens. It matches One-D-Piece at 64 tokens and achieves better reconstruction at 128 and 256 tokens. Compared with DetailFlow, NesTok provides substantially lower rFID at 32--128 tokens, although DetailFlow achieves better reconstruction at 256 tokens. While FlexTok and Semanticist obtain lower rFID at smaller token lengths, NesTok supports reconstruction across token lengths using a ViT decoder, without iterative diffusion or flow-based decoding.

\begin{table}[h]
    \centering
    
    \begin{tabular}{l|cccc}
    \toprule
    Method & 32 Tokens & 64 Tokens & 128 Tokens & 256 Tokens \\
    \midrule
    One-D-Piece & 3.23 & 2.10 & 1.42 & 1.08 \\
    ReTok-S-B     & 4.72 & 2.66 & 1.56 & 1.01 \\
    FlexTok d18-d28 & 1.45 & 1.37 & 1.20 & 1.08 \\
    Semanticist (DiT-XL) & 1.40 & 1.07 & 0.86 & 0.72 \\
    DetailFlow & 64.59 & 21.61 & 6.24 & 0.77 \\
    NesTok & 3.49 & 2.10 & 1.39 & 0.98 \\
    \bottomrule
    \end{tabular}
    \caption{Reconstruction performance (rFID$\downarrow$) at different token lengths on ImageNet $256\times256$.}
    \label{tab:rfid_different_tokens}

\end{table}

\begin{figure}[t]
    \centering
    \includegraphics[width=.99\linewidth]{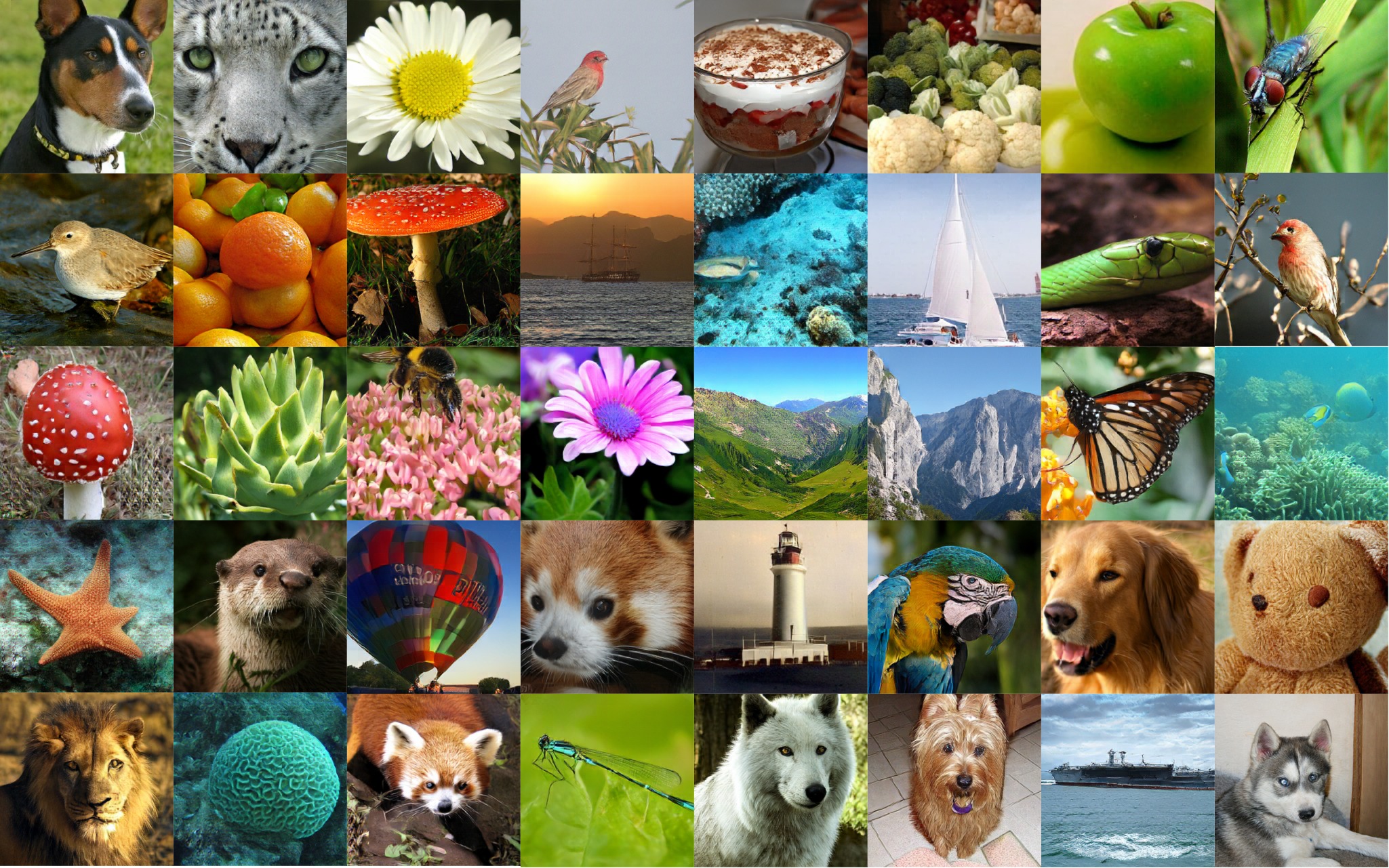}
    \caption{\textbf{Visualization of class-condition image generation on $256\times256$ resolution.} 
}
    \label{fig:imagenet_display}
    \vspace{-5pt}
\end{figure}

\textbf{More generation results. }
Tab.~\ref{tab:gfid_different_tokens} compares variable-length tokenizers with ViT decoders. NesTok-L achieves the lowest gFID among the compared methods at every evaluated token length, both with and without cfg. Using a single trained AR model, increasing the token length from 32 to 256 reduces its gFID from 4.82 to 2.08 without cfg and from 3.78 to 1.50 with cfg. Without cfg, One-D-Piece-L deteriorates at longer sequence lengths, while ReTok-L exhibits diminishing gains beyond 128 tokens. In contrast, NesTok-L continues to improve. These results demonstrate strong generation performance across token lengths and support the effective use of additional tokens for progressive refinement. Fig.~\ref{fig:imagenet_display} provides more visualization results on class-condition image generation with 256 tokens.

\begin{table}[t!]
    \centering
    
    \resizebox{\textwidth}{!}{
    \begin{tabular}{l|cccccccc}
    \toprule
    \multirow{2}{*}{Method} &
    \multicolumn{2}{c}{32 Tokens} &
    \multicolumn{2}{c}{64 Tokens} &
    \multicolumn{2}{c}{128 Tokens} &
    \multicolumn{2}{c}{256 Tokens} \\
    \cmidrule(lr){2-3}
    \cmidrule(lr){4-5}
    \cmidrule(lr){6-7}
    \cmidrule(lr){8-9}
    & w/o cfg & cfg
    & w/o cfg & cfg
    & w/o cfg & cfg
    & w/o cfg & cfg \\
    \midrule
    One-D-Piece-L
    & 8.30 & 5.27
    & 8.28 & 3.07
    & 10.91 & 2.56
    & 13.01 & 2.47 \\
    
    ReTok-L
    & 7.45 & 6.76
    & 4.69 & 4.54
    & 3.79 & 3.18
    & 3.78 & 2.66 \\

    DetailFlow-32
    & 67.59 & 60.34
    & 28.19 & 19.63
    & 12.88 & 6.77
    & 6.43 & 2.63 \\

    \midrule
    NesTok-L
    & \textbf{4.82} & \textbf{3.78}
    & \textbf{2.87} & \textbf{2.34}
    & \textbf{2.28} & \textbf{1.71}
    & \textbf{2.08} & \textbf{1.50} \\
    \bottomrule
    \end{tabular}}
    \caption{Comparison of generation FID (gFID$\downarrow$) at different token lengths on ImageNet $256\times256$.}
    \label{tab:gfid_different_tokens}

\end{table}

%% file: iclr2027_conference.bib
@article{ho2020denoising,
  title={Denoising diffusion probabilistic models},
  author={Ho, Jonathan and Jain, Ajay and Abbeel, Pieter},
  journal={Advances in neural information processing systems},
  volume={33},
  pages={6840--6851},
  year={2020}
}

@ARTICLE{autoprog,
  author={Li, Changlin and Zhang, Jiawei and Lin, Sihao and Yang, Zongxin and Liang, Junwei and Liang, Xiaodan and Chang, Xiaojun},
  journal={IEEE Transactions on Pattern Analysis and Machine Intelligence}, 
  title={Efficient Training of Large Vision Models via Advanced Automated Progressive Learning}, 
  year={2026},
  volume={48},
  number={8},
  pages={8798-8812},
  doi={10.1109/TPAMI.2026.3673336}}

@article{he2025rear,
  title={Rear: Rethinking visual autoregressive models via generator-tokenizer consistency regularization},
  author={He, Qiyuan and Li, Yicong and Ye, Haotian and Wang, Jinghao and Liao, Xinyao and Heng, Pheng-Ann and Ermon, Stefano and Zou, James and Yao, Angela},
  journal={arXiv preprint arXiv:2510.04450},
  year={2025}
}

@InProceedings{Lu_2026_CVPR,
    author    = {Lu, Jiasen and Song, Liangchen and Xu, Mingze and Ahn, Byeongjoo and Wang, Yanjun and Chen, Chen and Dehghan, Afshin and Yang, Yinfei},
    title     = {AToken: A Unified Tokenizer for Vision},
    booktitle = {Proceedings of the IEEE/CVF Conference on Computer Vision and Pattern Recognition (CVPR)},
    month     = {June},
    year      = {2026},
    pages     = {28701-28711}
}

@inproceedings{tatitok,
  title={Democratizing text-to-image masked generative models with compact text-aware one-dimensional tokens},
  author={Kim, Dongwon and He, Ju and Yu, Qihang and Yang, Chenglin and Shen, Xiaohui and Kwak, Suha and Chen, Liang-Chieh},
  booktitle={2025 IEEE/CVF International Conference on Computer Vision (ICCV)},
  pages={18442--18452},
  year={2025},
  organization={IEEE}
}

@inproceedings{xiong2025gigatok,
  title={Gigatok: Scaling visual tokenizers to 3 billion parameters for autoregressive image generation},
  author={Xiong, Tianwei and Liew, Jun Hao and Huang, Zilong and Feng, Jiashi and Liu, Xihui},
  booktitle={2025 IEEE/CVF International Conference on Computer Vision (ICCV)},
  pages={18770--18780},
  year={2025},
  organization={IEEE}
}

@inproceedings{D_ar,
  title={D-ar: Diffusion via autoregressive models},
  author={Gao, Ziteng and Shou, Mike Zheng},
  booktitle={International Conference on Learning Representations},
  volume={2026},
  pages={68013--68032},
  year={2026}
}

@inproceedings{huang2025spectralar,
  title={Spectralar: Spectral autoregressive visual generation},
  author={Huang, Yuanhui and Chen, Weiliang and Zheng, Wenzhao and Duan, Yueqi and Zhou, Jie and Lu, Jiwen},
  booktitle={2025 IEEE/CVF International Conference on Computer Vision (ICCV)},
  pages={15842--15852},
  year={2025},
  organization={IEEE}
}

@inproceedings{lee2022autoregressive,
  title={Autoregressive image generation using residual quantization},
  author={Lee, Doyup and Kim, Chiheon and Kim, Saehoon and Cho, Minsu and Han, Wook-Shin},
  booktitle={2022 IEEE/CVF conference on computer vision and pattern recognition (CVPR)},
  pages={11513--11522},
  year={2022},
  organization={IEEE}
}

@article{sun2024autoregressive,
  title={Autoregressive model beats diffusion: Llama for scalable image generation},
  author={Sun, Peize and Jiang, Yi and Chen, Shoufa and Zhang, Shilong and Peng, Bingyue and Luo, Ping and Yuan, Zehuan},
  journal={arXiv preprint arXiv:2406.06525},
  year={2024}
}

@article{vq-vae,
  title={Neural discrete representation learning},
  author={Van Den Oord, Aaron and Vinyals, Oriol and others},
  journal={Advances in neural information processing systems},
  volume={30},
  year={2017}
}

@article{vq-vae2,
  title={Generating diverse high-fidelity images with vq-vae-2},
  author={Razavi, Ali and Van den Oord, Aaron and Vinyals, Oriol},
  journal={Advances in neural information processing systems},
  volume={32},
  year={2019}
}

@inproceedings{vq-gan,
  title={Taming transformers for high-resolution image synthesis},
  author={Esser, Patrick and Rombach, Robin and Ommer, Bjorn},
  booktitle={Proceedings of the IEEE/CVF conference on computer vision and pattern recognition},
  pages={12873--12883},
  year={2021}
}

@article{titok,
  title={An image is worth 32 tokens for reconstruction and generation},
  author={Yu, Qihang and Weber, Mark and Deng, Xueqing and Shen, Xiaohui and Cremers, Daniel and Chen, Liang-Chieh},
  journal={Advances in Neural Information Processing Systems},
  volume={37},
  pages={128940--128966},
  year={2024}
}

@article{vae,
  title={Auto-encoding variational bayes},
  author={Kingma, Diederik P and Welling, Max},
  journal={arXiv preprint arXiv:1312.6114},
  year={2013}
}

@article{retok,
  title={Improving Flexible Image Tokenizers for Autoregressive Image Generation},
  author={Fu, Zixuan and Guo, Lanqing and Wang, Chong and Song, Binbin and Liu, Ding and Wen, Bihan},
  journal={arXiv preprint arXiv:2601.01535},
  year={2026}
}

@InProceedings{catok,
    author    = {Chen, Yitong and Wu, Zuxuan and Qiu, Xipeng and Jiang, Yu-Gang},
    title     = {CaTok: Taming Mean Flows for One-Dimensional Causal Image Tokenization},
    booktitle = {Proceedings of the IEEE/CVF Conference on Computer Vision and Pattern Recognition (CVPR)},
    month     = {June},
    year      = {2026},
    pages     = {23161-23171}
}

@article{chu2026end,
  title={End-to-end autoregressive image generation with 1d semantic tokenizer},
  author={Chu, Wenda and Zhang, Bingliang and Han, Jiaqi and Li, Yizhuo and Yang, Linjie and Yue, Yisong and Guo, Qiushan},
  journal={arXiv preprint arXiv:2605.00503},
  year={2026}
}

@inproceedings{chang2022maskgit,
  title={Maskgit: Masked generative image transformer},
  author={Chang, Huiwen and Zhang, Han and Jiang, Lu and Liu, Ce and Freeman, William T},
  booktitle={2022 IEEE/CVF Conference on Computer Vision and Pattern Recognition (CVPR)},
  pages={11305--11315},
  year={2022},
  organization={IEEE}
}

@inproceedings{magvit,
  title={Magvit: Masked generative video transformer},
  author={Yu, Lijun and Cheng, Yong and Sohn, Kihyuk and Lezama, Jos{\'e} and Zhang, Han and Chang, Huiwen and Hauptmann, Alexander G and Yang, Ming-Hsuan and Hao, Yuan and Essa, Irfan and others},
  booktitle={2023 IEEE/CVF Conference on Computer Vision and Pattern Recognition (CVPR)},
  pages={10459--10469},
  year={2023},
  organization={IEEE}
}

@inproceedings{magvit-v2,
  title={Language model beats diffusion-tokenizer is key to visual generation},
  author={Yu, Lijun and Lezama, Jos{\'e} and Gundavarapu, Nitesh Bharadwaj and Versari, Luca and Sohn, Kihyuk and Minnen, David and Cheng, Yong and Gupta, Agrim and Gu, Xiuye and Hauptmann, Alexander G and others},
  booktitle={International Conference on Learning Representations},
  volume={2024},
  pages={765--783},
  year={2024}
}

@inproceedings{softvq,
  title={Softvq-vae: Efficient 1-dimensional continuous tokenizer},
  author={Chen, Hao and Wang, Ze and Li, Xiang and Sun, Ximeng and Chen, Fangyi and Liu, Jiang and Wang, Jindong and Raj, Bhiksha and Liu, Zicheng and Barsoum, Emad},
  booktitle={2025 IEEE/CVF Conference on Computer Vision and Pattern Recognition (CVPR)},
  pages={28358--28370},
  year={2025},
  organization={IEEE}
}

@article{semtok,
  title={Semantic One-Dimensional Tokenizer for Image Reconstruction and Generation},
  author={Qu, Yunpeng and Zhang, Kaidong and Ding, Yukang and Chen, Ying and Wang, Jian},
  journal={arXiv preprint arXiv:2603.16373},
  year={2026}
}

@inproceedings{fsq,
  title={Finite scalar quantization: Vq-vae made simple},
  author={Mentzer, Fabian and Minnen, David and Agustsson, Eirikur and Tschannen, Michael},
  booktitle={International Conference on Learning Representations},
  volume={2024},
  pages={51772--51783},
  year={2024}
}

@inproceedings{rq-vae,
  title={Autoregressive image generation using residual quantization},
  author={Lee, Doyup and Kim, Chiheon and Kim, Saehoon and Cho, Minsu and Han, Wook-Shin},
  booktitle={2022 IEEE/CVF conference on computer vision and pattern recognition (CVPR)},
  pages={11513--11522},
  year={2022},
  organization={IEEE}
}

@inproceedings{ibq,
  title={Scalable image tokenization with index backpropagation quantization},
  author={Shi, Fengyuan and Luo, Zhuoyan and Ge, Yixiao and Yang, Yujiu and Shan, Ying and Wang, Limin},
  booktitle={2025 IEEE/CVF International Conference on Computer Vision (ICCV)},
  pages={16037--16046},
  year={2025},
  organization={IEEE}
}

@article{llamagen,
  title={Autoregressive model beats diffusion: Llama for scalable image generation},
  author={Sun, Peize and Jiang, Yi and Chen, Shoufa and Zhang, Shilong and Peng, Bingyue and Luo, Ping and Yuan, Zehuan},
  journal={arXiv preprint arXiv:2406.06525},
  year={2024}
}

@article{mar,
  title={Autoregressive image generation without vector quantization},
  author={Li, Tianhong and Tian, Yonglong and Li, He and Deng, Mingyang and He, Kaiming},
  journal={Advances in Neural Information Processing Systems},
  volume={37},
  pages={56424--56445},
  year={2024}
}

@article{mactok,
  title={MacTok: Robust Continuous Tokenization for Image Generation},
  author={Zeng, Hengyu and Gao, Xin and Li, Guanghao and Yan, Yuxiang and Ruan, Jiaoyang and Ma, Junpeng and Wang, Haoyu Albert and Pu, Jian},
  journal={arXiv preprint arXiv:2603.29634},
  year={2026}
}

@inproceedings{flextok,
  title={Flextok: Resampling images into 1d token sequences of flexible length},
  author={Bachmann, Roman and Allardice, Jesse and Mizrahi, David and Fini, Enrico and Kar, O{\u{g}}uzhan Fatih and Amirloo, Elmira and El-Nouby, Alaaeldin and Zamir, Amir and Dehghan, Afshin},
  booktitle={Forty-second International Conference on Machine Learning},
  year={2025}
}

@inproceedings{semanticist,
  title={“Principal Components” Enable A New Language of Images},
  author={Wen, Xin and Zhao, Bingchen and Elezi, Ismail and Deng, Jiankang and Qi, Xiaojuan},
  booktitle={2025 IEEE/CVF International Conference on Computer Vision (ICCV)},
  pages={16641--16651},
  year={2025},
  organization={IEEE}
}

@article{var,
  title={Visual autoregressive modeling: Scalable image generation via next-scale prediction},
  author={Tian, Keyu and Jiang, Yi and Yuan, Zehuan and Peng, Bingyue and Wang, Liwei},
  journal={Advances in neural information processing systems},
  volume={37},
  pages={84839--84865},
  year={2024}
}

@inproceedings{randar,
  title={Randar: Decoder-only autoregressive visual generation in random orders},
  author={Pang, Ziqi and Zhang, Tianyuan and Luan, Fujun and Man, Yunze and Tan, Hao and Zhang, Kai and Freeman, William T and Wang, Yu-Xiong},
  booktitle={2025 IEEE/CVF Conference on Computer Vision and Pattern Recognition (CVPR)},
  pages={45--55},
  year={2025},
  organization={IEEE}
}

@article{one-d-piece,
  title={One-d-piece: Image tokenizer meets quality-controllable compression},
  author={Miwa, Keita and Sasaki, Kento and Arai, Hidehisa and Takahashi, Tsubasa and Yamaguchi, Yu},
  journal={arXiv preprint arXiv:2501.10064},
  year={2025}
}

@inproceedings{rar,
  title={Randomized autoregressive visual generation},
  author={Yu, Qihang and He, Ju and Deng, Xueqing and Shen, Xiaohui and Chen, Liang-Chieh},
  booktitle={2025 IEEE/CVF International Conference on Computer Vision (ICCV)},
  pages={18431--18441},
  year={2025},
  organization={IEEE}
}

@article{yu2024image,
  title={An image is worth 32 tokens for reconstruction and generation},
  author={Yu, Qihang and Weber, Mark and Deng, Xueqing and Shen, Xiaohui and Cremers, Daniel and Chen, Liang-Chieh},
  journal={Advances in Neural Information Processing Systems},
  volume={37},
  pages={128940--128966},
  year={2024}
}

@inproceedings{zheng2023online,
  title={Online clustered codebook},
  author={Zheng, Chuanxia and Vedaldi, Andrea},
  booktitle={2023 IEEE/CVF International Conference on Computer Vision (ICCV)},
  pages={22741--22750},
  year={2023},
  organization={IEEE}
}

@article{su2024roformer,
  title={Roformer: Enhanced transformer with rotary position embedding},
  author={Su, Jianlin and Ahmed, Murtadha and Lu, Yu and Pan, Shengfeng and Bo, Wen and Liu, Yunfeng},
  journal={Neurocomputing},
  volume={568},
  pages={127063},
  year={2024},
  publisher={Elsevier}
}

@article{dhariwal2021diffusion,
  title={Diffusion models beat gans on image synthesis},
  author={Dhariwal, Prafulla and Nichol, Alexander},
  journal={Advances in neural information processing systems},
  volume={34},
  pages={8780--8794},
  year={2021}
}

@article{team2024chameleon,
  title={Chameleon: Mixed-modal early-fusion foundation models},
  author={Team, Chameleon},
  journal={arXiv preprint arXiv:2405.09818},
  year={2024}
}

@article{zhang2019root,
  title={Root mean square layer normalization},
  author={Zhang, Biao and Sennrich, Rico},
  journal={Advances in neural information processing systems},
  volume={32},
  year={2019}
}

@inproceedings{lightingdit,
  title={Reconstruction vs. generation: Taming optimization dilemma in latent diffusion models},
  author={Yao, Jingfeng and Yang, Bin and Wang, Xinggang},
  booktitle={2025 IEEE/CVF Conference on Computer Vision and Pattern Recognition (CVPR)},
  pages={15703--15712},
  year={2025},
  organization={IEEE}
}

@inproceedings{alit,
  title={Adaptive length image tokenization via recurrent allocation},
  author={Duggal, Shivam and Isola, Phillip and Torralba, Antonio and Freeman, William T},
  booktitle={First Workshop on Scalable Optimization for Efficient and Adaptive Foundation Models},
  year={2025}
}

@inproceedings{wang2025visual,
  title={Visual lexicon: Rich image features in language space},
  author={Wang, XuDong and Zhou, Xingyi and Fathi, Alireza and Darrell, Trevor and Schmid, Cordelia},
  booktitle={2025 IEEE/CVF Conference on Computer Vision and Pattern Recognition (CVPR)},
  pages={19736--19747},
  year={2025},
  organization={IEEE}
}

@inproceedings{dit,
  title={Scalable diffusion models with transformers},
  author={Peebles, William and Xie, Saining},
  booktitle={2023 IEEE/CVF International Conference on Computer Vision (ICCV)},
  pages={4172--4182},
  year={2023},
  organization={IEEE}
}

@article{repa,
  title={Representation alignment for generation: Training diffusion transformers is easier than you think},
  author={Yu, Sihyun and Kwak, Sangkyung and Jang, Huiwon and Jeong, Jongheon and Huang, Jonathan and Shin, Jinwoo and Xie, Saining},
  journal={arXiv preprint arXiv:2410.06940},
  year={2024}
}

@article{detailflow,
  title={Detailflow: 1d coarse-to-fine autoregressive image generation via next-detail prediction},
  author={Liu, Yiheng and Qu, Liao and Zhang, Huichao and Wang, Xu and Jiang, Yi and Gao, Yiming and Ye, Hu and Li, Xian and Wang, Shuai and Du, Daniel K and others},
  journal={arXiv preprint arXiv:2505.21473},
  year={2025}
}

@article{restok,
  title={ResTok: Learning Hierarchical Residuals in 1D Visual Tokenizers for Autoregressive Image Generation},
  author={Zhang, Xu and Da, Cheng and Yang, Huan and Gai, Kun and Lu, Ming and Ma, Zhan},
  journal={arXiv preprint arXiv:2601.03955},
  year={2026}
}

@article{alitok,
  title={Alitok: Towards sequence modeling alignment between tokenizer and autoregressive model},
  author={Wu, Pingyu and Zhu, Kai and Liu, Yu and Tang, Longxiang and Yang, Jian and Peng, Yansong and Zhai, Wei and Cao, Yang and Zha, Zheng-Jun},
  journal={arXiv e-prints},
  pages={arXiv--2506},
  year={2025}
}
